\documentclass[10pt,twocolumn,letterpaper]{article}

\usepackage{cvpr}              % To produce the CAMERA-READY version
\usepackage{graphicx}
\usepackage{amsmath}
\usepackage{amssymb}
\usepackage{booktabs}
\usepackage{comment}

\usepackage[pagebackref,breaklinks,colorlinks]{hyperref}

\usepackage[capitalize]{cleveref}
\crefname{section}{Sec.}{Secs.}
\Crefname{section}{Section}{Sections}
\Crefname{table}{Table}{Tables}
\crefname{table}{Tab.}{Tabs.}

\def\confName{Arxiv}
\def\confYear{2022}

\newcommand{\OURS}{MVLayoutNet}

\begin{document}

\title{\OURS{}: 3D layout reconstruction with multi-view panoramas}

\author{
Zhihua Hu$^{1,2}$ \qquad Bo Duan$^{1}$  \qquad Yanfeng Zhang$^{1}$ \qquad Mingwei Sun$^{1,2}$ \qquad Jingwei Huang$^{1}$
\vspace{0.2cm} \\ 
$^{1}$Riemann Lab, Huawei Technologies \qquad $^{2}$Wuhan University
\vspace{-0.3cm}
}

\maketitle

\begin{abstract}

We present \OURS{}, an end-to-end network for holistic 3D reconstruction from multi-view panoramas.
Our core contribution is to seamlessly combine learned monocular layout estimation and multi-view stereo (MVS) for accurate layout reconstruction in both 3D and image space.
We jointly train a layout module to produce an initial layout and a novel MVS module to obtain accurate layout geometry.
Unlike standard MVSNet~\cite{MVSNet}, our MVS module takes a newly-proposed layout cost volume, which aggregates multi-view costs at the same depth layer into corresponding layout elements. We additionally provide an attention-based scheme that guides the MVS module to focus on structural regions.
Such a design considers both local pixel-level costs and global holistic information for better reconstruction.
Experiments show that our method outperforms state-of-the-arts in terms of depth rmse by 21.7\% and 20.6\% on the 2D-3D-S~\cite{2d3ds} and ZInD~\cite{ZInD} datasets.
Finally, our method leads to coherent layout geometry that enables the reconstruction of an entire scene.
\end{abstract}

\section{Introduction}
\begin{figure}
\includegraphics[width=\linewidth]{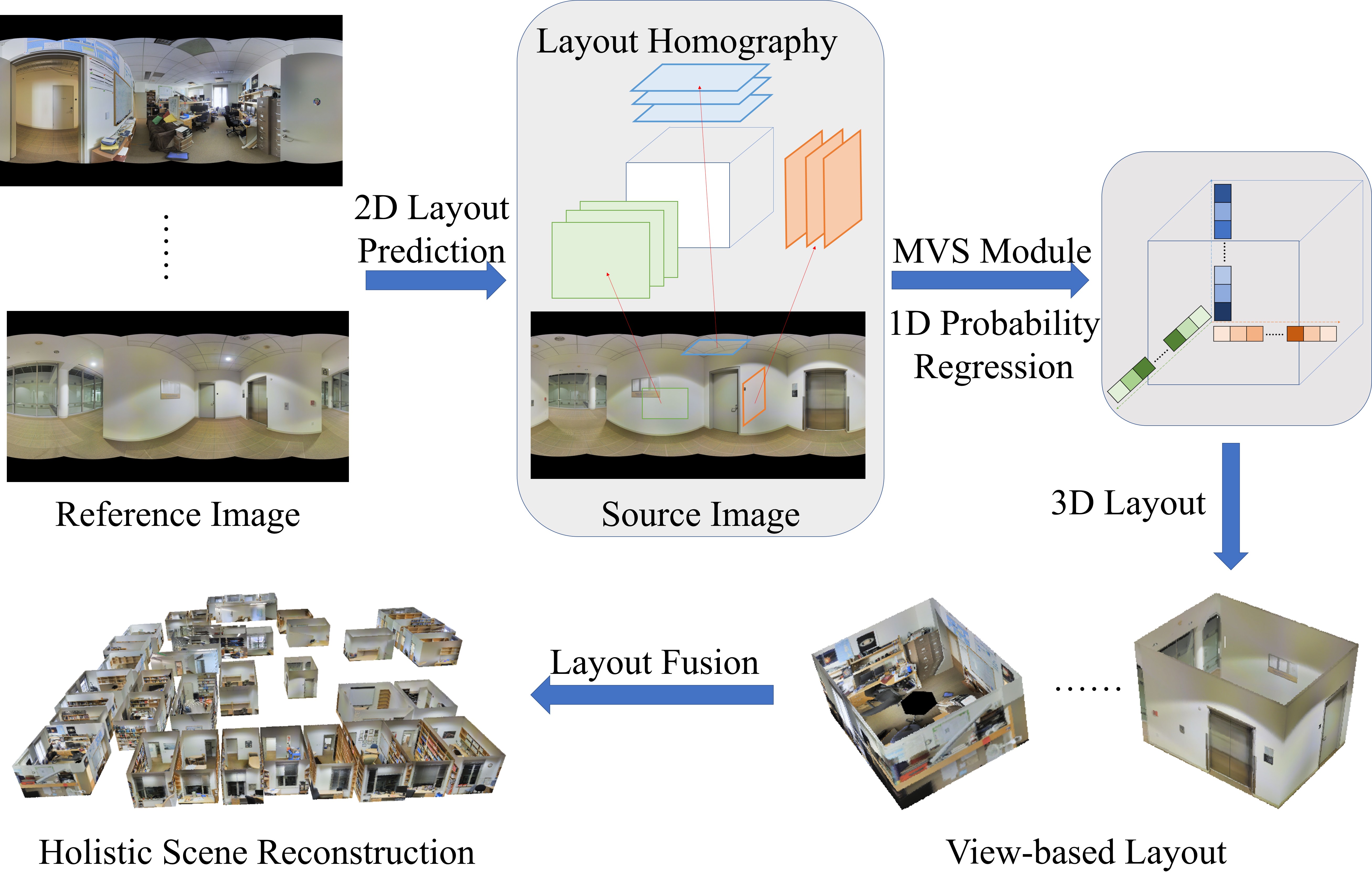}
\caption{\OURS{} starts from a 2D layout prediction and introduces an MVS module that takes a layout cost volume and provides a one-dimensional probability aggregation to predict layout depths. MVS-based 3D layouts provide coherent global holistic reconstruction.}
\label{fig:teaser}
\end{figure}

% Background
Holistic reconstruction of the 3D environment is a fundamental problem in the computer vision and graphics community. It aims to produce abstract and clean models to support various applications, including simulation, gaming, and virtual/augmented reality. While many devices are suitable for 3D reconstruction, commodity RGB cameras are still one of the most popular choices since they are usually available to common people.
Multiview stereo (MVS) is the standard technique for image-based 3D reconstruction and has been studied for decades, including traditional methods~\cite{SGM,PMVS} and learning-based methods~\cite{MVSNet,yi2020PVAMVSNET}. Results produced by traditional approaches often lack integrity caused by smooth texture, reflections, and transparent materials. Learning-based methods~\cite{MVSNet,yi2020PVAMVSNET} aims to alleviate these problems by building deep feature volumes, but their results still suffer from noises and incompletion. While various post-processing methods filter the point cloud from the MVS by surface reconstruction~\cite{kazhdan2006poisson,kazhdan2013screened}, primitive fitting~\cite{sfpn,Huang_2021_ICCV} or shape generation~\cite{deepsdf,llig}, the produced quality is still not guaranteed to satisfy user demands.

Recently, a new set of approaches directly estimate holistic layouts from monocular panoramas~\cite{horizonnet,lednet,panocontext,layoutnet}. Since the layout representation enforces a well-formulated room shape, the methods guarantee the integrity of the produced reconstruction. However, a monocular panorama can only cover a limited space, which prevents these methods from reconstructions of large-scale scenes or complex environments with occlusions. Further, since a monocular view lacks multiview information, the 3D network can only learn from limited RGB information, leading to inaccurate 3D estimation in various environments.

% Philosophy
Intuitively, an ideal solution would be to fully take advantage of both layout estimation and MVS to address the integrity and accuracy limitations. Some works tried to consider both techniques: \cite{LeeYPY17,Cabral14} assumes reconstructed geometry from MVS or range data and estimates layouts. \cite{Micusik09} improves MVS from local planarity constraints that are relevant to layout information. However, neither of them compensates limitations of each technique from the other. To our knowledge, this fundamental problem has not been seriously answered.
Our goal is to seamlessly combine both MVS and monocular layouts for accurate holistic reconstruction.

We present \OURS{} as an end-to-end network to tackle the problem as shown in Figure~\ref{fig:teaser}. Our network mainly consists of a layout module followed by an MVS module.
Our layout module serves to predict an initial layout and can be implemented by existing methods~\cite{horizonnet,lednet,panocontext,layoutnet}.
Our key contribution is the MVS module built upon estimated 2D layouts. We first build a novel layout cost volume by computing feature costs via newly-proposed layout homography transforms.
To tackle a cluttered environment where structures are partially occluded, we incorporate explicit semantic information and an attention branch to highlight contributions of layout cost volumes corresponding to structural regions.
The layout cost volume adjusted by the learned attention is passed through our MVS module to predict a single depth value for each layout element.
The key idea of this module is to perform cost aggregation and compress information into a one-dimensional probability map for each layout element.
It leads to a highly abstract representation of volume information and enforces the network to analyze the overall structure rather than untrustworthy local regions. More importantly, the probability regression is performed in 1D, which narrows the space for depth search and leads to more robust and generalizable layout reconstruction.

Comparisons on the 2D-3D-Semantics Dataset (2D-3D-S) \cite{2d3ds} and Zillow Indoor Dataset (ZInD) \cite{ZInD} show that we significantly improve the geometric accuracy of the 3D layout reconstruction. Specifically, we achieve a significant improvement of 21.7\% and 20.6\% in terms of depth root mean square error (RMSE) on the 2D-3D-S and ZInD datasets. We observe that our MVS module leads to more robust depth estimation in challenging environments where layout boundaries are with smooth textures. This advantage still holds when comparing our layout-based MVS with the standard pixel-level MVS Network~\cite{MVSNet}. From our ablation studies (Sec.~\ref{sec:exp-ablate}), wrong predictions from \cite{MVSNet} at textureless regions seriously decrease the quality of the layout. This problem is resolved in our design which focuses on overall structures. Experiments also show that specific semantic filtering with weakly supervised self-attention can further improve the accuracy of the 3D layout.

To sum up, our contributions are as follows:
\vspace{-0.1in}
\begin{itemize}
    \item We present an end-to-end network that seamlessly combines layout estimation and multiview stereo for large-scale scene layout reconstruction.
    \vspace{-0.1in}
    \item We propose a layout cost volume with a novel MVS module that performs 1D probability regression on abstract layout elements.
    \vspace{-0.1in}
    \item We use explicit semantics and weakly supervised self-attention to enforce structural analysis.
    \vspace{-0.1in}
    \item Experiments highlight contributions of our key designs and significant quality improvement from the state-of-the-arts. 
\end{itemize}

\section{Related Works}

\paragraph{Multi-view stereo}
Multi-view stereo (MVS) has been a hot topic in the vision community for decades. The core problem of MVS is how to find the reliable correspondences.

Traditional methods designed many kinds of photo-consistency measures (e.g. Normalized Cross Correlation) and reconstruction schemes (e.g. Semi-Global Matching \cite{SGM} and Patch Match \cite{PMVS}). These methods can achieve very good results when the textures are rich. However, it is hard to recover reliable dense correspondences for traditional MVS methods due to the commonly existing texture-less areas in indoor scenes.

With the development of Convolutional Neural Network (CNN), many learning based MVS methods have been proposed. For example, MVSNet \cite{MVSNet} estimates the depth maps from multi-view images with learned features, differentiable homography and 3D cost volume regularization. PVA-MVSNet \cite{yi2020PVAMVSNET} utilizes the pyramid network and self-adaptive view aggregation to improve the completeness and accuracy of the reconstructions. 
However, reliable dense correspondences still cannot be obtained in the texture-less areas.

\paragraph{Room layout estimation}
Generally, there are two ways to reconstruct the room layouts from images: one is firstly recovering the point clouds with MVS and then estimating the layout from the point clouds while the others treat the layout estimation task as a segmentation or feature (plane boundaries or room corners) extraction problem and estimate the layout directly from the perspective or panoramic image. 

There are few works focused on the former pipelines. 
The authors of \cite{Cabral14} propose a method to reconstruct piecewise planar and compact floorplans from multiple images through a structure classification technique and solve the shortest path problem on a specially crafted graph. However, it is hard to obtain satisfied estimation results when there are not enough reliable 3D points, especially with sparse multi-view images.

Instead, many researchers explored the latter pipelines which estimate the room layout directly from the monocular image. Since the 2D layout in image space can be represented with lines or corner points, the earlier layout estimation methods focus on extracting lines and corner points of images \cite{tra1,tra2,tra3} and formulate the layout estimation problem as an optimization problem \cite{op1,op2}. Recently, many learning based methods have been proposed and show better estimation results compared to traditional methods \cite{im2cad,roomnet}. 

Starting from the work of \cite{panocontext}, researchers have focused on estimating layouts from monocular panorama due to the wide FOV of it. Similar to layout estimation from perspective images, geometric and semantic cues \cite{geo1,geo2,geo3,geo4,geo5} have been explored to better estimate the room layouts. Also, methods based on CNN gradually dominate the layout estimation from monocular panorama with the better performance \cite{layoutnet,horizonnet,AtlantaNet,lednet,HoHoNet}.

Room layouts estimated with monocular image generally have no scale information. To obtain the 3D layout, assumptions about the camera height are often made, e.g. the camera heights are set to 1.6 m \cite{horizonnet}. However, the real camera height can be arbitrary which will lead to imprecise and inconsistent 3D layouts. In addition, the accurate layout boundaries are hard to be estimated from the monocular image due to the occlusions from the clutters such as tables and chairs. 
\par
%proposed
In this paper, we present a learning based 3D layout reconstruction method with multi-view panoramas to address these issues. We combine the advantages of MVS and monocular layout estimation to achieve accurate and coherent 3D layout reconstruction. We leverage the layout estimated from monocular panorama and directly reconstruct the layout elements instead of point or pixel with multi-view panoramas. Furthermore, semantics from images and self-attention mechanisms have been incorporated to achieve better results with the existence of clutters.

\section{Approach}
\label{sec:approach}
\subsection{Overview}
The input to our problem is a set of calibrated panoramas $\mathbf{I}_i$ with given camera extrinsic $\mathbf{T}_i$. Our goal is to reconstruct the global 3D layout $\mathcal{L}^G$, and our system design borrows both ideas from layout estimation and multi-view stereo (MVS). We first estimate a 2D layout $\mathcal{L}_i$  from each panorama $\mathbf{I}_i$ through a 2D network (Section~\ref{sec:approach-2d}). Similar to MVS, we treat each view $i$ as a reference view where several related source views are aggregated as a cost volume. Section~\ref{sec:approach-3d} introduces the formulation of our layout cost volume $\mathcal{C}_i$ given $\mathcal{L}_i$ as a basis, and our MVS module which reconstruct the 3D layout $\mathcal{\hat{L}}_i$ from $\mathcal{C}_i$. The 3D layout can further be improved with semantic information and weakly-supervised self-attention (Section~\ref{sec:approach-attention}). All network modules are trained in an end-to-end manner. Finally, we aggregate 3D layouts $\mathcal{\hat{L}}_i$ to get $\mathcal{L}^G$ via layout fusion (Section~\ref{sec:approach-fusion}).

\begin{figure*}
\includegraphics[width=\linewidth]{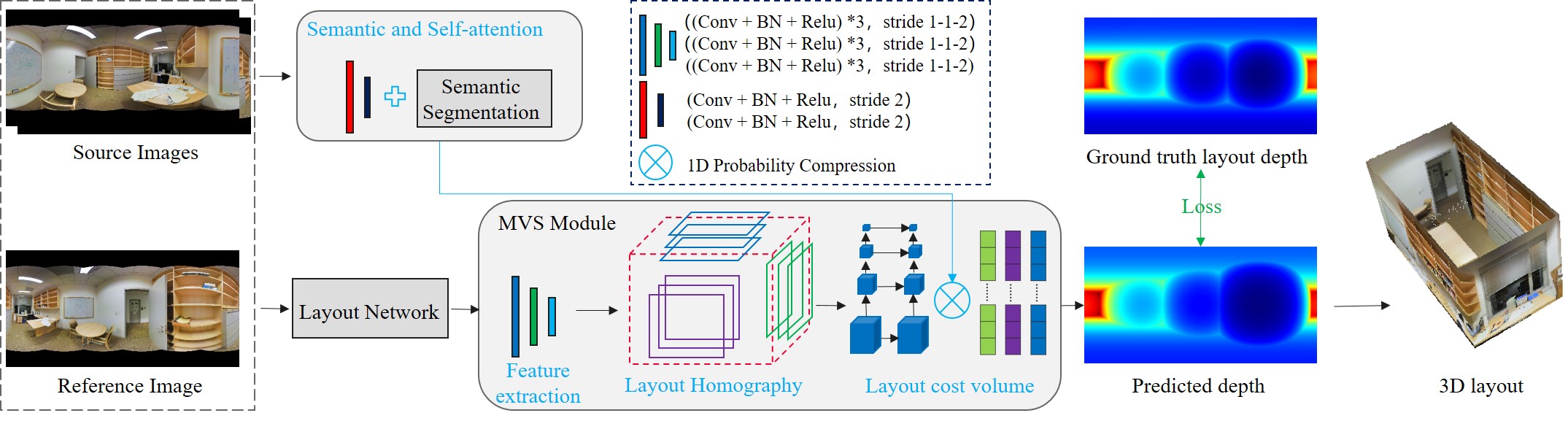}
\caption{\OURS{} pipeline. We estimate the 2D layout of a reference view via a layout network. The layout is fed into the MVS module, where we build layout cost volume with layout homography and compress the 3D volume into a 1D layout probability map to determine the 3D layout. Additional semantics and self-attention mechanisms are incorporated to reduce negative influences caused by object occlusions in cluttered scenes.}
\label{fig:pipeline}
\end{figure*}

\subsection{2D Layout Estimation}
\label{sec:approach-2d}
A 2D Layout $\mathcal{L}=\{\mathcal{E}_j\}$ can be viewed as a piecewise planar model covering the whole panorama, where each plane is a layout element  $\mathcal{E}_{j}=<\mathbf{P}_j,\mathbf{o}_j>$ consisting of a planar region $\mathbf{P}_j$ in the image space with a plane orientation $\mathbf{o}_j$. We make reasonable assumptions that $\mathbf{P}_j$ is either horizontal or vertical, and the y-axis of the panorama is aligned with the up-vector in the scene. Such a layout can be estimated from existing layout networks~\cite{layoutnet,horizonnet,AtlantaNet,lednet,HoHoNet} and our network design can accept any of them, as shown in Figure~\ref{fig:pipeline}. Although these methods provide additional 3D layout prediction, they assume connectivity and fixed camera height and need only predict relative scales. We find that such a prediction is still inaccurate, and argue that these strong assumptions prevent more general environment topology reconstruction. Therefore, our design does not impose any of these constraints and only borrows their capability to predict layouts in 2D, thereby extensible to complex settings in the future.

We use~\cite{horizonnet} as our layout module, which produces 2D layout corner points, their relative depths, and edges connecting point pairs among them. $\mathbf{P}_j$ can be determined by applying a breadth-first search in the image space until reaching the layout boundaries specified by edges. Since orientation is not relevant to the absolute geometry scale, we can compute layout element orientation $\mathbf{o}_j$ as the normal of the polygon formed by layout corners with predicted relative depths. The 2D layout loss can be computed with the corner points $\mathbf{C}_i$ and edges $\mathbf{B}_i$ as Equation~\ref{eq:2dloss}:
\begin{equation}
\mathbf{L}^{\mathcal{L}}(\mathbf{C}_i,\mathbf{B}_i)=\sum\text{CrossEntropy}(\mathbf{C}_i,\mathbf{C}_i^*)+|\mathbf{B}_i-\mathbf{B}_i^*|
\label{eq:2dloss}
\end{equation}
Where the $\mathbf{C}_i^*$ and $\mathbf{B}_i^*$ are the corner points and edges of the ground truth layout.
\subsection{3D Layout Reconstruction}
\label{sec:approach-3d}
Our MVS module consists of three components. We apply layout homography given 2D layout prediction of the reference view and multiple source panorama views. Feature volumes are aggregated and compressed into a layout cost volume, which is passed through a probability estimation module to get a one-dimensional probability map. Layout depth is finally determined by integrating depth intervals with probabilities.

\paragraph{Layout Homography}
We treat each panorama $\mathbf{I}_i$ as the reference view with a set of associated source views $\mathbf{R}_i$. Details of view selection are described in Section~\ref{sec:exp-comparison}.
Standard cost volume is constructed by enumerating several depth planes orthogonal to the reference view, efficiently warping source views to the reference camera space via homography, and measuring differences based on local patches~\cite{PMVS} or deep features~\cite{MVSNet}. However, it does not precisely model the physical world where the orientation of an ideally-aligned patch should be specified by the surface normal. Therefore, we determine the cost for each pixel $\mathbf{p}$ and a depth $d$ in the reference view using homography specified by more accurate layout element orientation instead of the vanilla optical axis, described in Equation~\ref{eq:homography}. We slightly abuse the term ``depth'' here as the distance between the layout element plane and the camera location.
\begin{equation}
\mathbf{H}_s(\mathbf{p},d) = \mathbf{K}_s\cdot\mathbf{R}_s\cdot(\mathbf{I}-\frac{(\mathbf{t}_i-\mathbf{t}_s)\cdot \mathbf{o}_{i,j}^T}{d})\cdot\mathbf{K}_i^T\cdot\mathbf{R}_i^T
\label{eq:homography}
\end{equation}
where $s$ is any source view in $\mathbf{R}_i$ related to reference view $i$. $\mathbf{o}_{i,j}$ is determined by the layout element from $\mathcal{L}_i$ whose region covers pixel $\mathbf{p}$.

\begin{figure}
\centering
\begin{minipage}{0.96\linewidth}
\centering
\includegraphics[width=\linewidth]{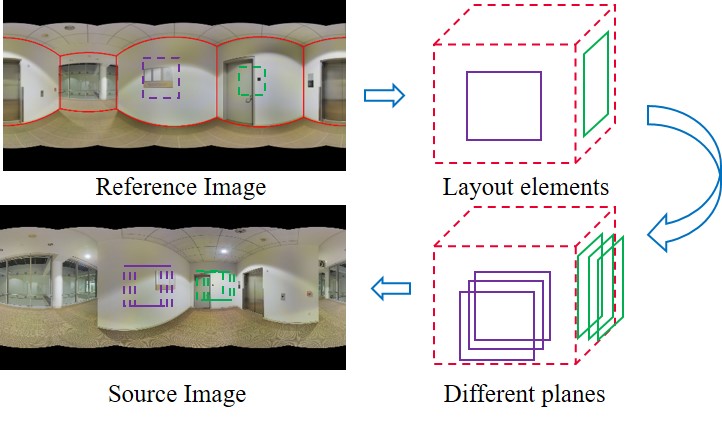}
(a) Layout homography
\end{minipage}
\begin{minipage}{0.96\linewidth}
\centering
\includegraphics[width=\linewidth]{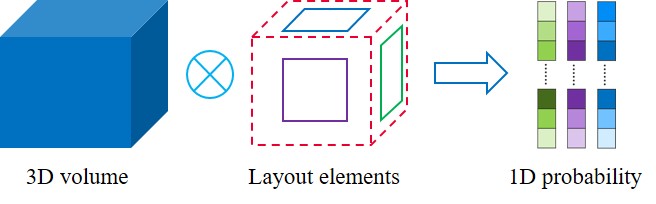}
(b) 1D probability compression
\end{minipage}
\caption{(a) We apply homography specified by orientations of layout elements rather than commonly used optical axis. (b) We compress 3D layout cost volume into a 1D probability map and regress a single depth for each layout element.}% and enforce structure analysis guided by a confidence map from semantics and self-attention.}
\label{fig:homography}
\end{figure}

Although local homography depends on pixel locations, layout regions are often continuous and clustered. Therefore, we need only enumerate all layout element orientation $\mathbf{o}_{i,j}$, apply homography transformation to regions in the source views which overlaps with the corresponding layout element in the reference view. The overall computation cost is almost the same as a single homography applied to the whole image.
Figure~\ref{fig:homography}(a) illustrates how we apply homography from different planes. Regions in reference views overlapping with different layout elements are highlighted with different colors and applied with separate layout homography given different orientations.

\paragraph{Layout cost volume}
Since our goal is to describe the overall geometry of the layout, we compress the probability volume to describe the layout element feature. First, we build a deep feature volume in the reference camera space for each source view similar to MVSNet~\cite{MVSNet} but via our layout homography. Then, the cost volume passes through a 3D-UNet to derive a volume probability $\text{Prob}(\mathbf{p}_k,d)$, which is further compressed into a one-dimensional probability map $\text{Prob}(\mathcal{E}_{i,j},d)$ for each layout element $\mathcal{E}_{i,j}$ by aggregating information at the same depth layer as shown in the Figure~\ref{fig:homography}(b). Mathematical details are described in Equation~\ref{eq:probability}.
\begin{comment}
\begin{equation}
\mathcal{C}_{i,j}(d)=\sum_{\mathbf{p}_k\in \mathbf{P}_{i,j}} \frac{c_k}{|\mathbf{P}_{i,j}|}\cdot \frac{\sum_{s\in\mathbf{R}_i}(\mathbf{V}_s(\mathbf{p}_k,d) - \mathbf{\overline{V}}_s(\mathbf{p}_k,d))^2}{|\mathbf{R}_i|}
\label{eq:cost}
\end{equation}
\end{comment}

\begin{equation}
\text{Prob}(\mathcal{E}_{i,j},d)=\sum_{\mathbf{p}_k\in \mathbf{P}_{i,j}} \frac{c_k}{|\mathbf{P}_{i,j}|}\cdot \frac{\sum_{s\in\mathbf{R}_i}\text{Prob}(\mathbf{p}_k,d)}{|\mathbf{R}_i|}
\label{eq:probability}
\end{equation}
$c_k$ is a learned confidence to adjust the contribution of pixel $\mathbf{p}_k$ to the layout depth estimation, which is discussed in Section~\ref{sec:approach-attention}.

We regress the layout depth by aggregating the one-dimensional probability map in Equation~\ref{eq:regress}.
\begin{equation}
\mathcal{D}(\mathcal{E}_{i,j}) = \sum_{d=d_\text{min}}^{d_\text{max}} \text{Prob}(\mathcal{E}_{i,j},d)\cdot d
\label{eq:regress}
\end{equation}

During training, the regressed depth is supervised by ground truth layout depth $\mathcal{D}^*$ using L1 norm (Equation~\ref{eq:depth-loss}).
\begin{equation}
\mathbf{L}_{i,j}^{\mathcal{D}} = |\mathcal{D}(\mathcal{E}_{i,j})-\mathcal{D}^*(\mathcal{E}_{i,j})|
\label{eq:depth-loss}
\end{equation}

\subsection{Semantics and Self-attention}
\label{sec:approach-attention}
Since real environments are usually cluttered, layout structures are partially occluded by objects in panoramas. Therefore, it is beneficial to highlight contributions of useful regions and deemphasize the influence of object regions when reconstructing the 3D layout. We rely on confidence score $c_k$ (Equation~\ref{eq:probability}) to adjust contributions at the pixel level. We incorporate semantic information as explicit supervision and adopt a weakly-supervised self-attention module to learn such an adjustment.

In detail, we use HoHoNet \cite{HoHoNet} to extract the semantic label map $\{\mathbf{S}_i\}$ for all input panoramas $\{\mathbf{I}_i\}$. Each semantic label corresponds to a confidence $c^s$ that is learned or specified by the user. When layout semantics are explicitly defined, we straightforwardly set $c^s\in\{0,1\}$ depending on whether the semantic belongs to the layout. Otherwise, we learn a MLP layer that translates semantic features into $c^s\in[0,1]$. We discuss details of the choices in Section~\ref{sec:exp-ablate}.
In the meanwhile, we pass $\mathbf{I}_i$ through a self-attention module which is a four-layer CNN to determine a learned confidence $c^a\in[0,1]$ for each pixel. We multiply them to obtain the final contribution $c_k=c^a_k\cdot c^s_k$ by considering both first principles and learned statistics.

\subsection{Implementation}
\label{sec:approach-fusion}
During training, we optimize all aforementioned network parameters in an end-to-end manner by minimizing the loss $\mathbf{L}$ in Equation~\ref{eq:loss}.
\begin{equation}
\mathbf{L}=\sum_{i}\mathbf{L}^{\mathcal{L}}(\mathbf{C}_i,\mathbf{B}_i)+\sum_{i,j}\mathbf{L}^{\mathcal{D}}(\mathcal{E}_{i,j})
\label{eq:loss}
\end{equation}

At inference time, we extract a 3D layout for each view by combining the 2D layout from the layout module and depth prediction from the MVS module. Layouts from each view are finally aggregated into a global world. In our implementation, layout elements from different views are fused if they are close and visible to both views given a depth threshold (0.1m). This fusion step leads to a clean, unified, and holistic reconstruction of an entire scene.

\section{Experiments}
We compare our method with the state-of-the-art methods in Section~\ref{sec:exp-comparison}, where we show a clear advantage from absolute 3D accuracy and relative geometry coherency. Section~\ref{sec:exp-ablate} performs ablation studies and shows that our proposed ideas can dramatically improve the reconstruction quality. We demonstrate a large-scale 3D layout reconstruction in Section~\ref{sec:exp-result}.

\subsection{Comparison}
\label{sec:exp-comparison}
\paragraph{Baselines} Since our goal is to reconstruct 3D layouts of the scene, we aim to compare with the state-of-the-arts sharing the same task~\cite{horizonnet,HoHoNet}. To our knowledge, these works are the best candidates for comparison since there are no studies that combine MVS with network inference of the layout as we do.
Since absolute depth is not available from these methods, we apply a standard scale prediction proposed by~\cite{DulaNet}, denoted as ``SP'' in all following experiments.
We notice that it is also possible to build 3D layouts as a postprocessing step after multi-view stereo, which we will discuss in Section~\ref{sec:exp-ablate}. 

\paragraph{Datasets} We propose to use 2D-3D-S~\cite{2d3ds} and ZInD~\cite{ZInD} as commonly used or recently released datasets for experiments. In 2D-3D-S, all panorama images from the same room are mutually visible and grouped together as related views. We set 112 groups of images in scene 5 as test data and all other 401 groups as training data. For ZInD, images are grouped based on their associated floor ID. As a result, we obtain 7536 groups of images for training and 945 groups of images for testing. Matterport3D~\cite{Matterport3D} is another commonly used dataset. But it lacks camera poses for skybox images and thus is not appropriate for our multi-view setting.

\paragraph{Metrics} We propose several metrics to evaluate the quality of the 3D layout. The depth RMSE is the rooted mean square error of the estimated layout depth map and the ground truth layout depth map. It measures the mean error between the predicted layout and the ground truth. The scale error is the absolute difference between the estimated and real camera heights. It serves as an overall metric reflecting how close the predicted scale of environment is to the ground truth. Finally, it is crucial that layouts from different views are consistent in the 3D world, which is measured by coherency as the mean distance of pairs of 3D points extracted from corresponding pixels in the reference and the source images.

\begin{figure}
\centering
\begin{minipage}{0.96\linewidth}
\centering
\includegraphics[width=\linewidth]{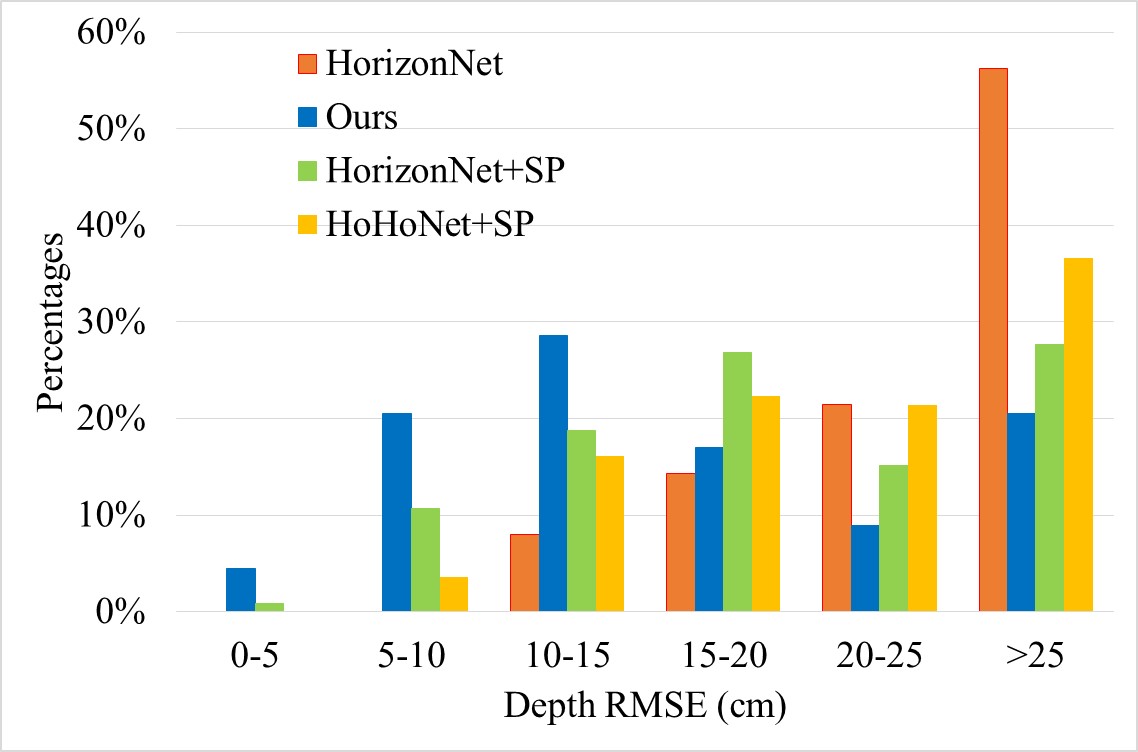}
\end{minipage}
\caption{The error distributions of different scenes according to the Depth RMSE on the 2D-3D-S dataset. We reconstruct more scenes with low depth RMSE and less scenes with high depth RMSE.}
\label{fig:histgram}
\end{figure}

\paragraph{Results} Table~\ref{tab:Accuracy_2d3ds} shows that our method outperforms the state-of-the-art methods in all metrics on the 2D-3D-S dataset~\cite{2d3ds}. Worth mentioning, we achieve an improvement of 21.7\% in terms of depth RMSE, and 50.0\% in the scale error. We observe that 2D-3D-S is captured at quite similar camera heights, which reduces the discrepancy of coherency scores among compared methods. 
%error distribution
As shown in the Figure~\ref{fig:histgram}, we can reconstruct more scenes in low depth RMSE which indicates the effectiveness of the proposed method. When the depth RMSE is larger than 25 cm, we have nearly 10\% less scenes which suggests that the our method is more robust.

\begin{figure*}
\centering
\begin{minipage}{0.24\linewidth}
\centering
\includegraphics[width=\linewidth]{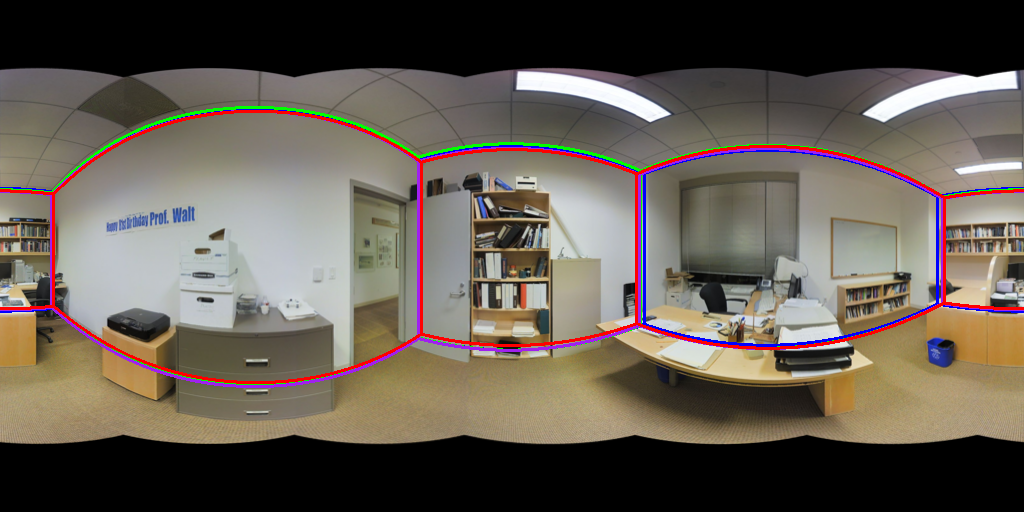}
\includegraphics[width=\linewidth]{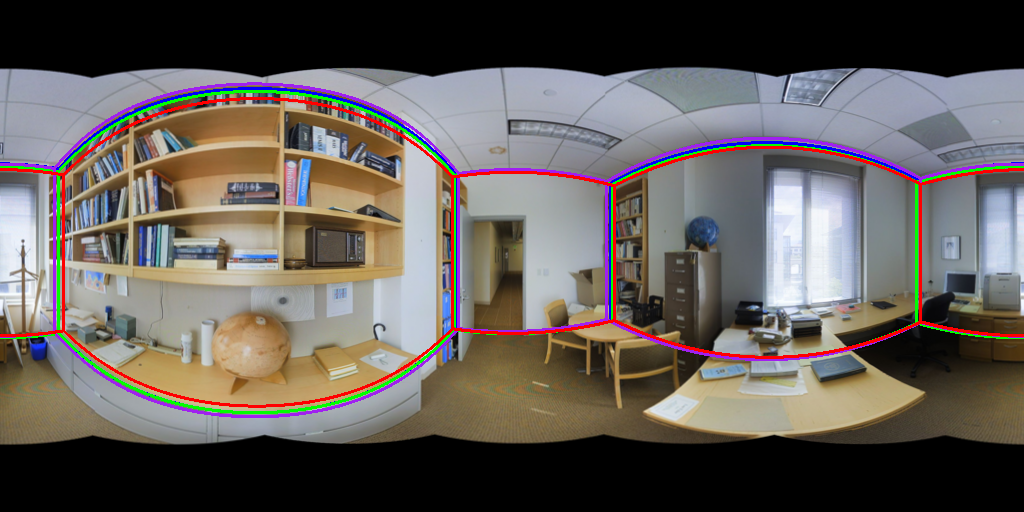}
\includegraphics[width=\linewidth]{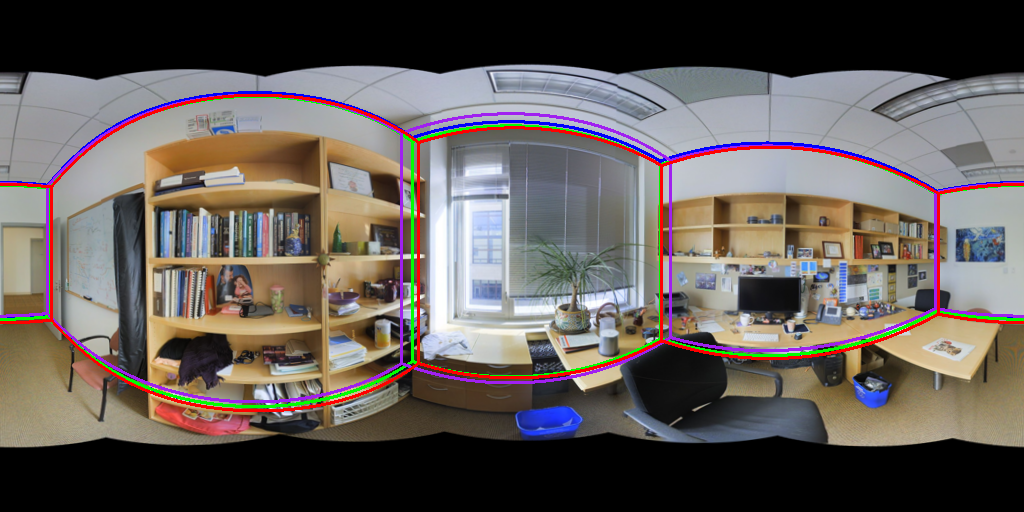}
\includegraphics[width=\linewidth]{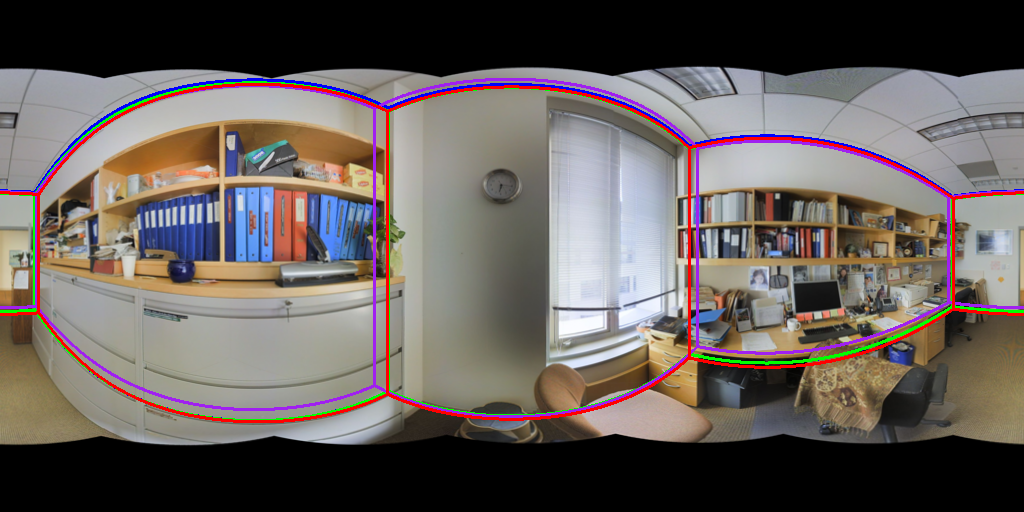}
(a) 2D layouts
\end{minipage}
\begin{minipage}{0.24\linewidth}
\centering
\includegraphics[width=\linewidth]{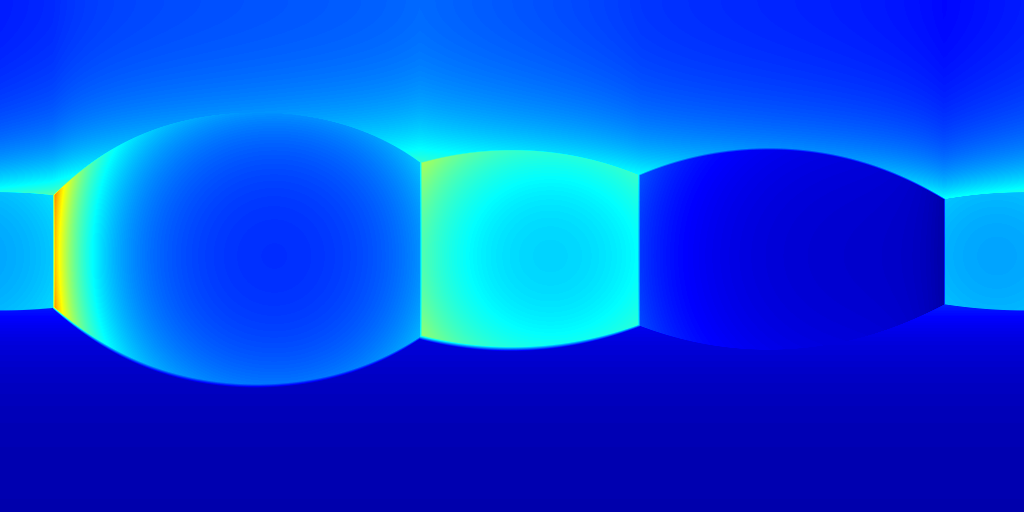}
\includegraphics[width=\linewidth]{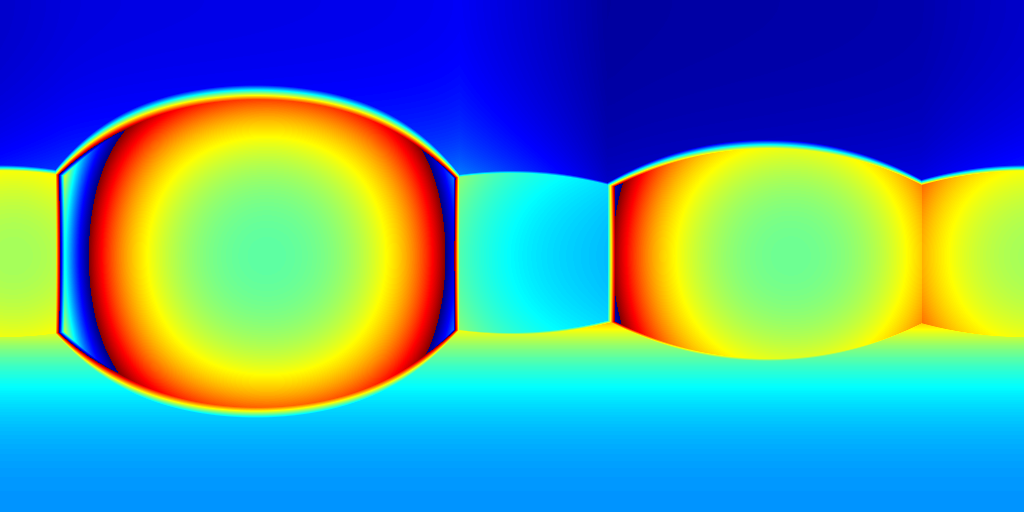}
\includegraphics[width=\linewidth]{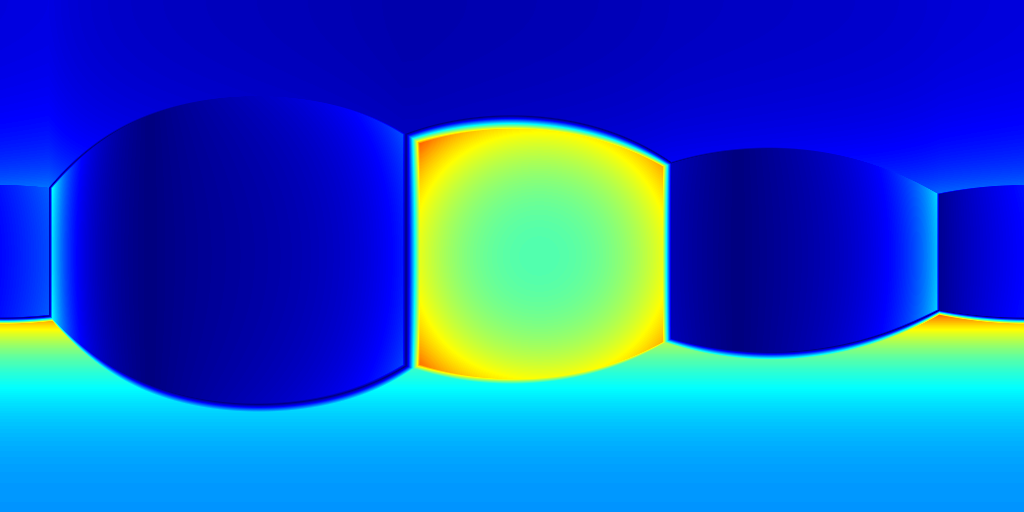}
\includegraphics[width=\linewidth]{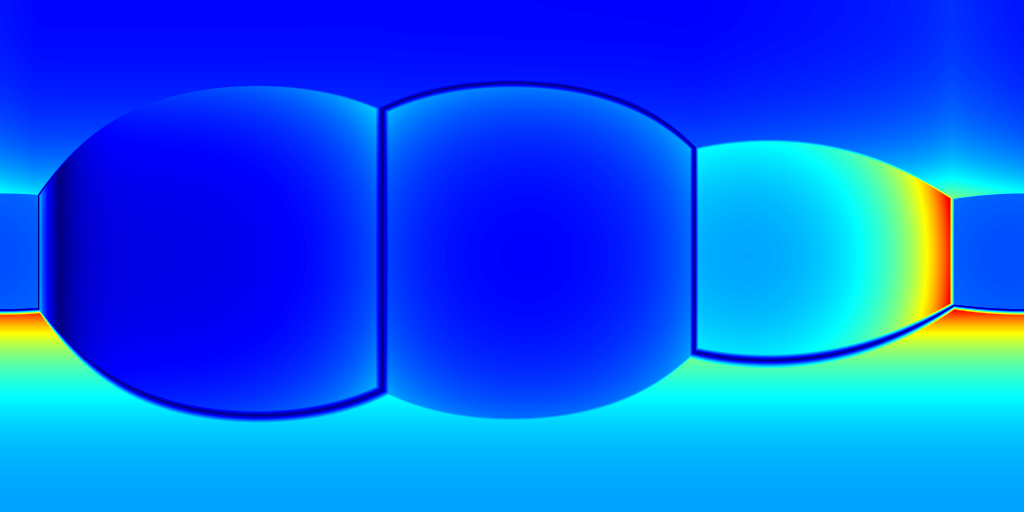}
(b) HoHoNet+SP
\end{minipage}
\begin{minipage}{0.24\linewidth}
\centering
\includegraphics[width=\linewidth]{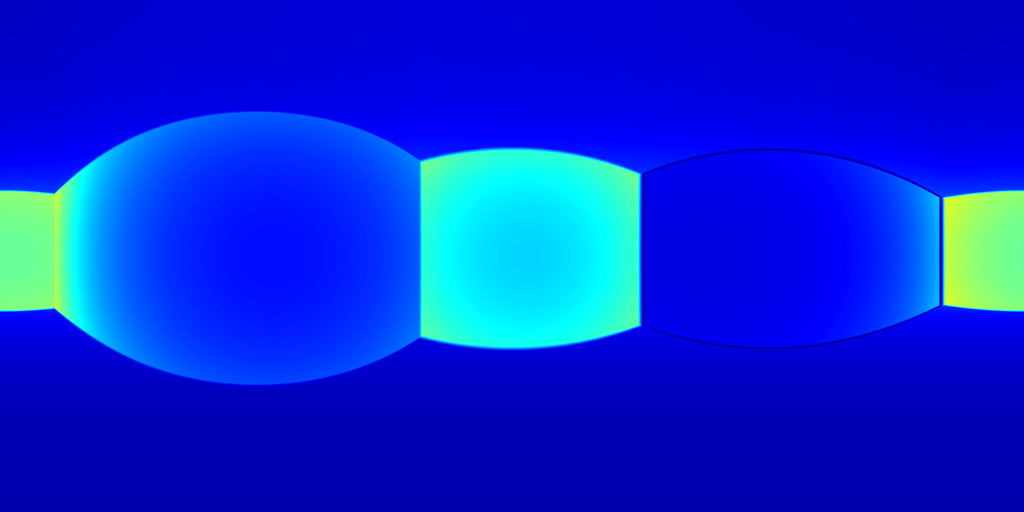}
\includegraphics[width=\linewidth]{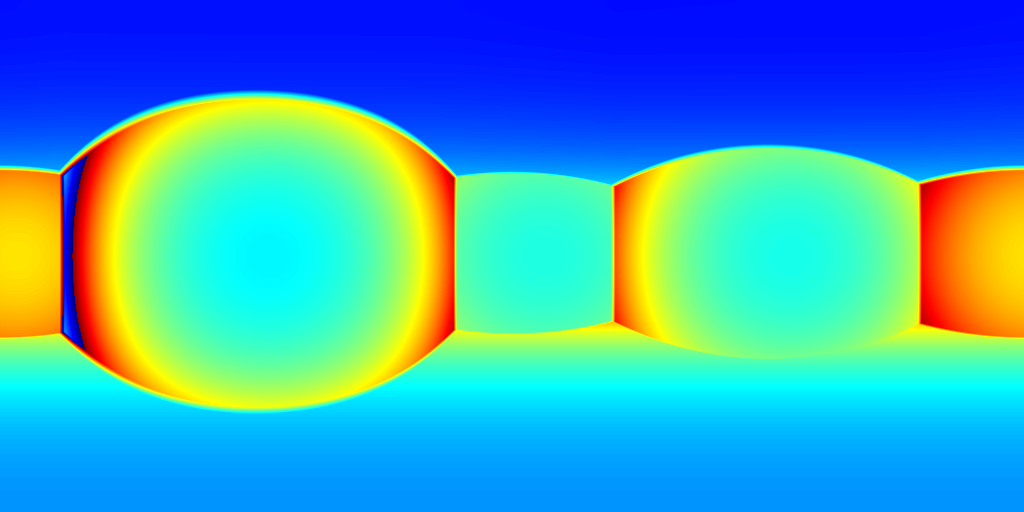}
\includegraphics[width=\linewidth]{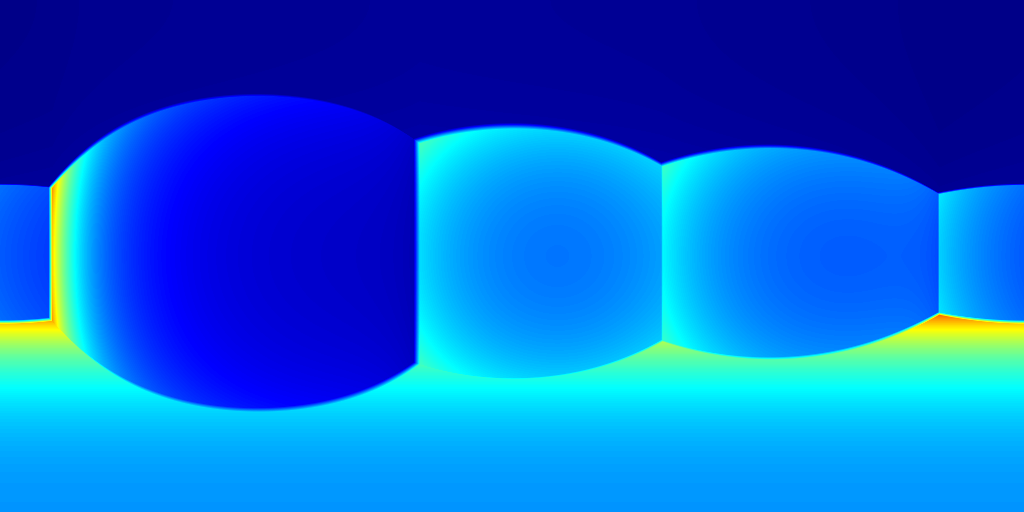}
\includegraphics[width=\linewidth]{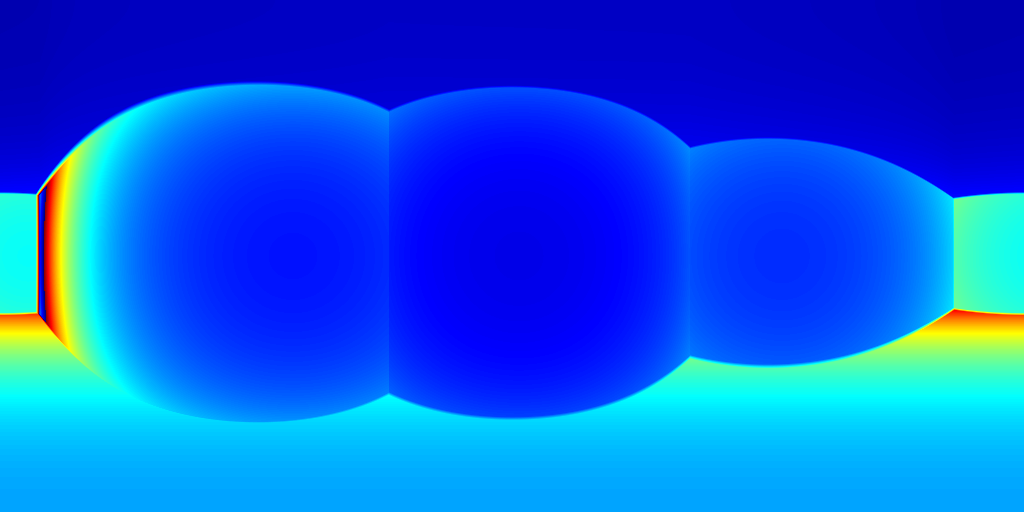}
(c) HorizonNet+SP
\end{minipage}
\begin{minipage}{0.24\linewidth}
\centering
\includegraphics[width=\linewidth]{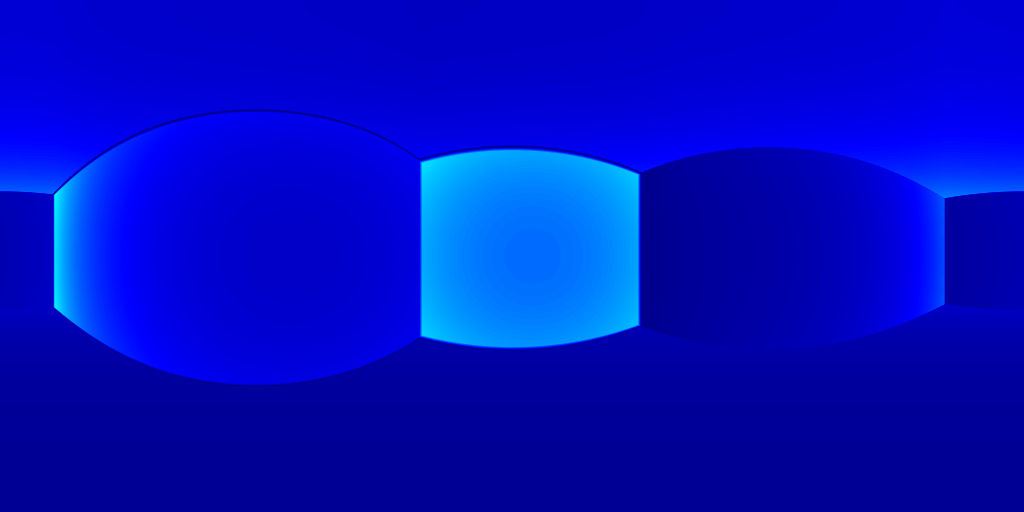}
\includegraphics[width=\linewidth]{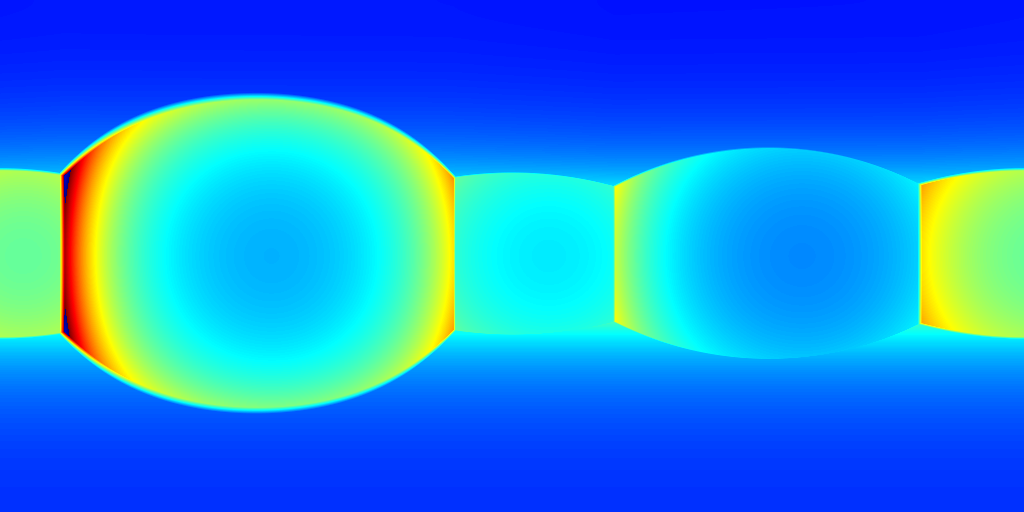}
\includegraphics[width=\linewidth]{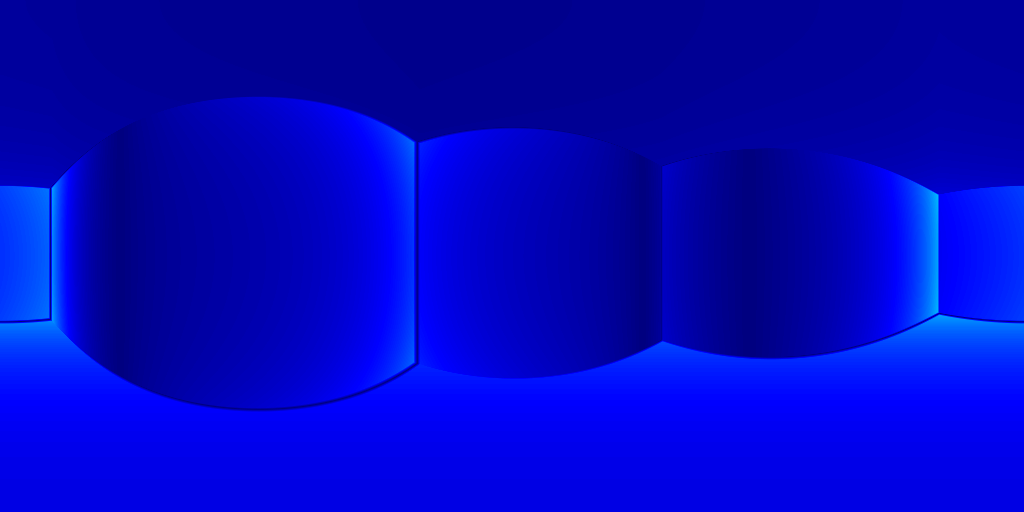}
\includegraphics[width=\linewidth]{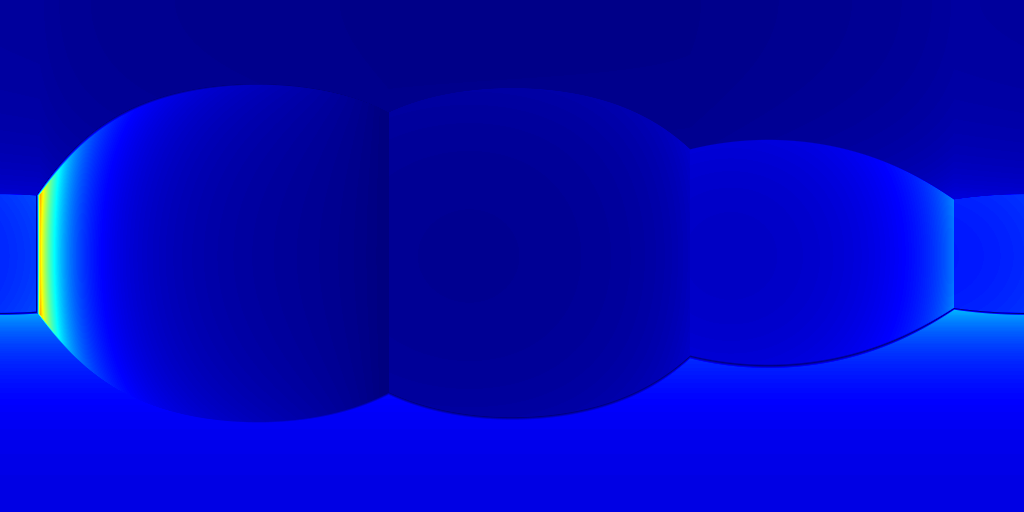}
(d) Ours
\end{minipage}
\caption{The qualitative results of layout boundary and depth error map on the 2D-3D-S dataset. The layout boundaries of Ground truth, HoHoNet, HorizonNet, and Ours are shown in red, purple, blue, and green colors respectively. The depth error maps are colored with the absolute difference of the estimated layout depth map and the ground truth layout depth map from blue (low) to red (high). Accurate layout boundaries and depth maps can be reconstructed in the scenes with furniture.}
\label{fig:2d3ds} 
\end{figure*}

We visualize the reconstructed layouts and the depth error maps for 2D-3D-S in Figure~\ref{fig:2d3ds}. Figure~\ref{fig:2d3ds}(a) plots the layout boundaries predicted from different methods, where the ground truth, HoHoNet~\cite{HoHoNet}, HorizonNet~\cite{horizonnet} and ours are visualized with red, purple, blue, and green colors respectively. As expected, the quality of layouts projected into the image space is quite similar among different methods. However, our improvement of 3D accuracy is significant as shown in Figure~\ref{fig:2d3ds}(b-d). 
\begin{table}
\centering
\small
\begin{tabular}{|c|c|c|c|}
\hline
2D-3D-S & Depth RMSE & Scale error & Coherency \\
\hline
HorizonNet & 0.28 & 0.17 & 0.11 \\
\hline
HorizonNet+SP & 0.23 & 0.10 & 0.10 \\
\hline
HoHoNet+SP & 0.26 & 0.10 & 0.12 \\
\hline
Ours & \textbf{0.18} & \textbf{0.05} & \textbf{0.09} \\
\hline
\end{tabular}
\caption{Evaluation for 3D layout reconstruction on 2D-3D-S dataset (in m). We achieve the best accuracy in terms of depth RMSE, scale error, and coherency.}
\label{tab:Accuracy_2d3ds}
\end{table}

\begin{figure*}
\centering
\begin{minipage}{0.24\linewidth}
\centering
\includegraphics[width=\linewidth]{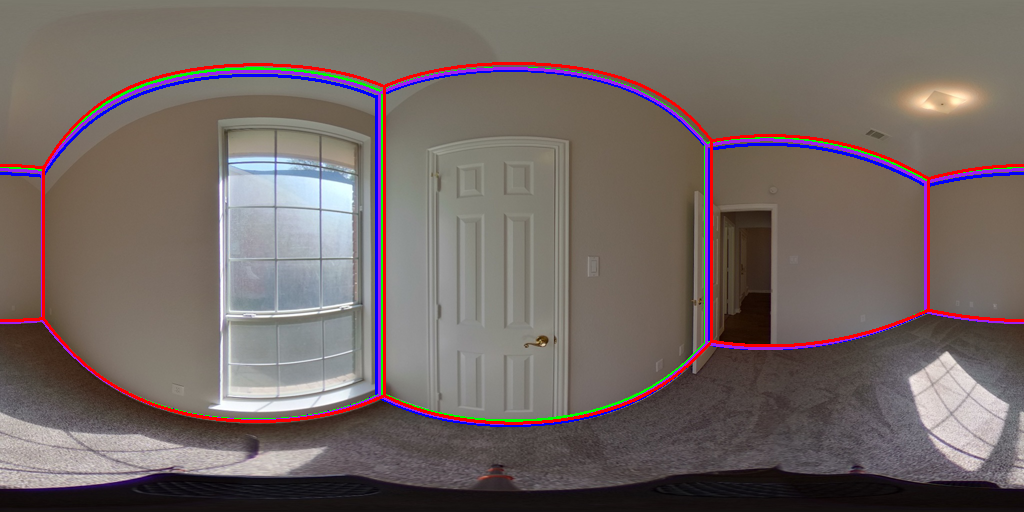}
\includegraphics[width=\linewidth]{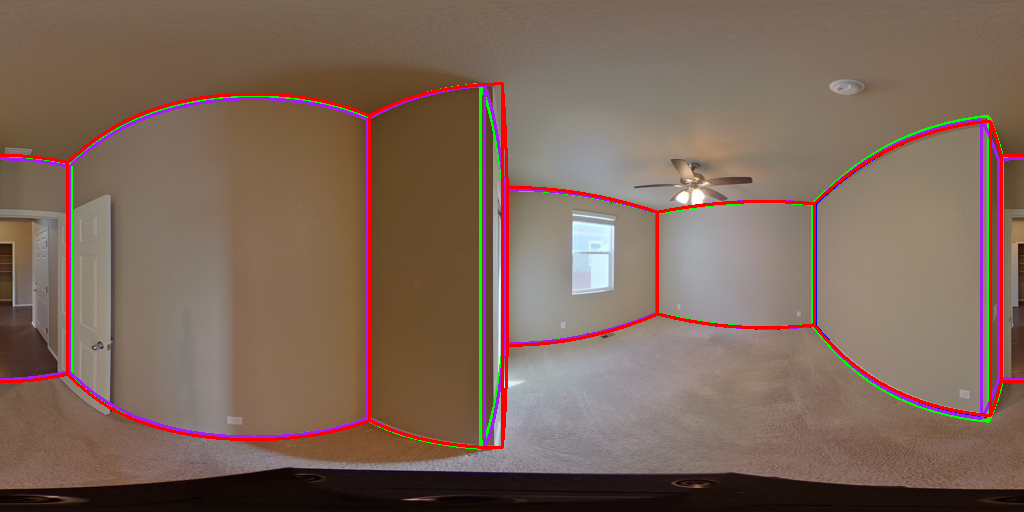}
\includegraphics[width=\linewidth]{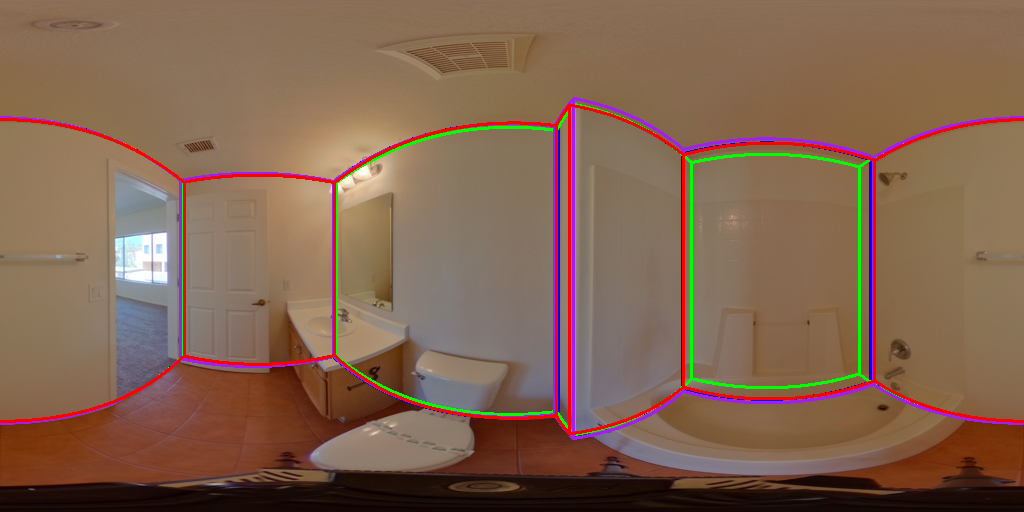}
\includegraphics[width=\linewidth]{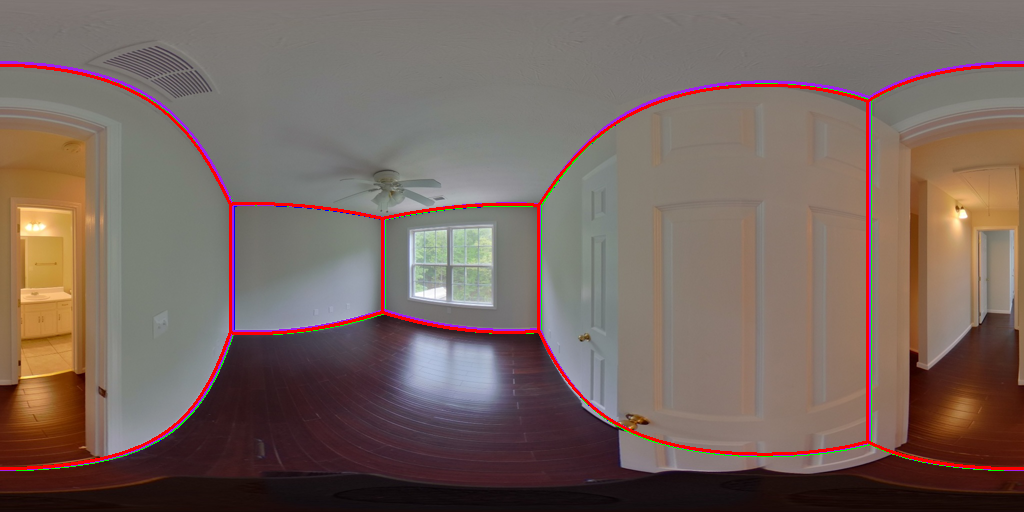}
(a) 2D layouts
\end{minipage}
\begin{minipage}{0.24\linewidth}
\centering
\includegraphics[width=\linewidth]{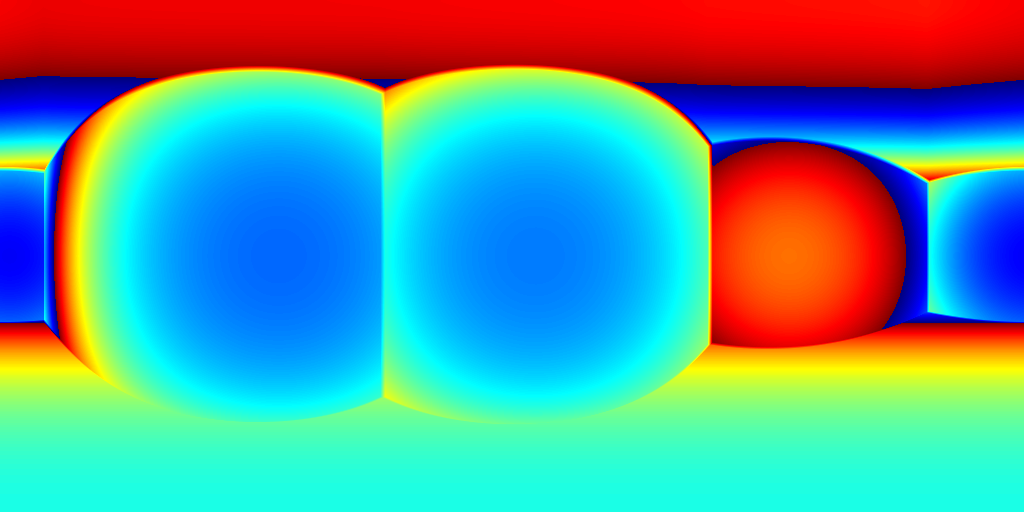}
\includegraphics[width=\linewidth]{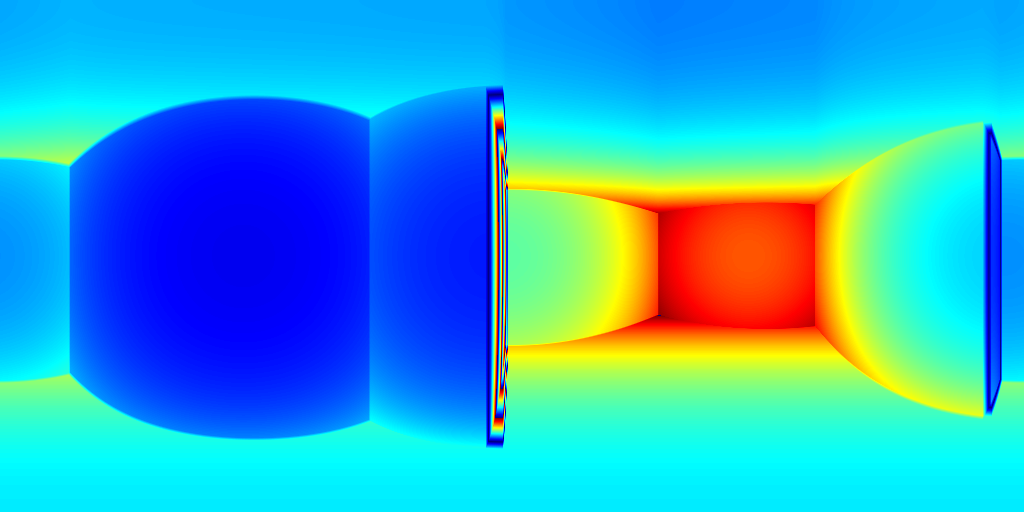}
\includegraphics[width=\linewidth]{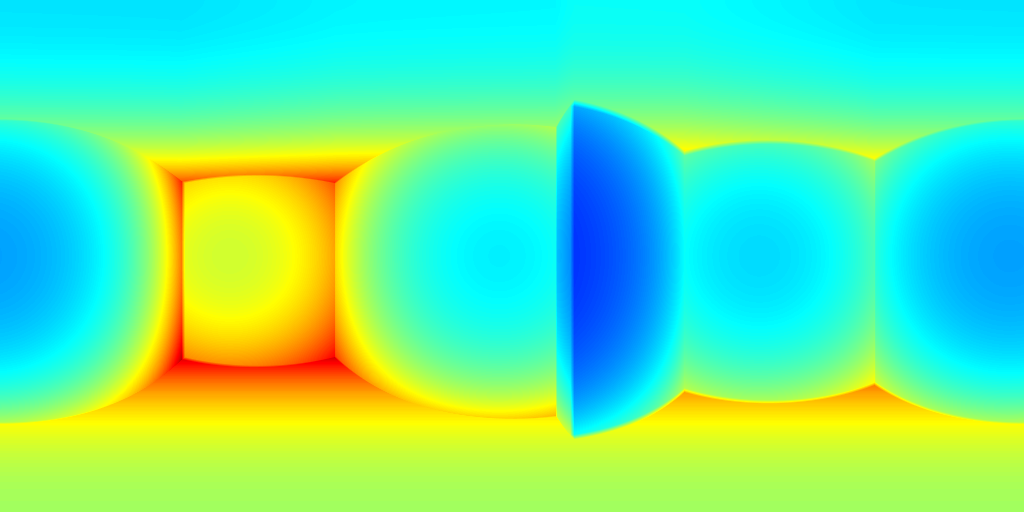}
\includegraphics[width=\linewidth]{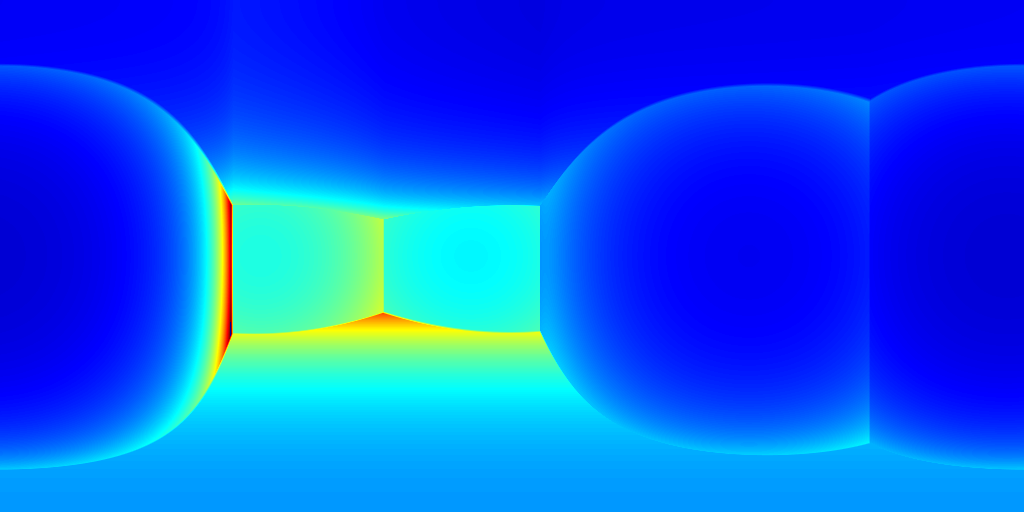}
(b) HoHoNet+SP
\end{minipage}
\begin{minipage}{0.24\linewidth}
\centering
\includegraphics[width=\linewidth]{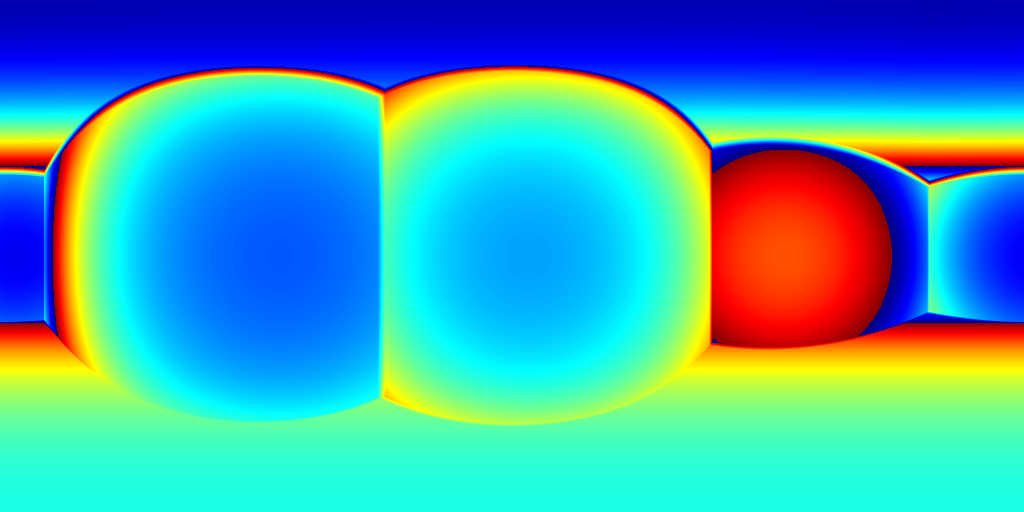}
\includegraphics[width=\linewidth]{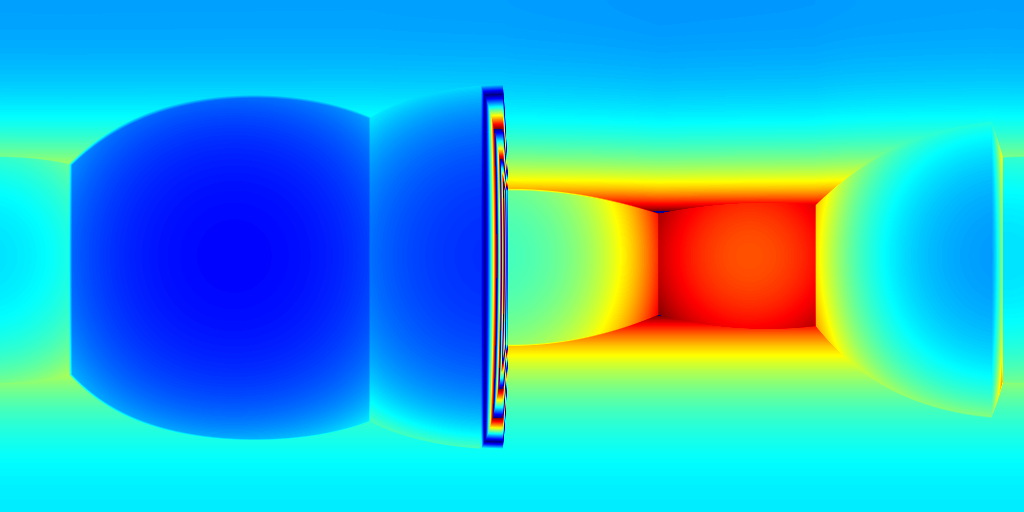}
\includegraphics[width=\linewidth]{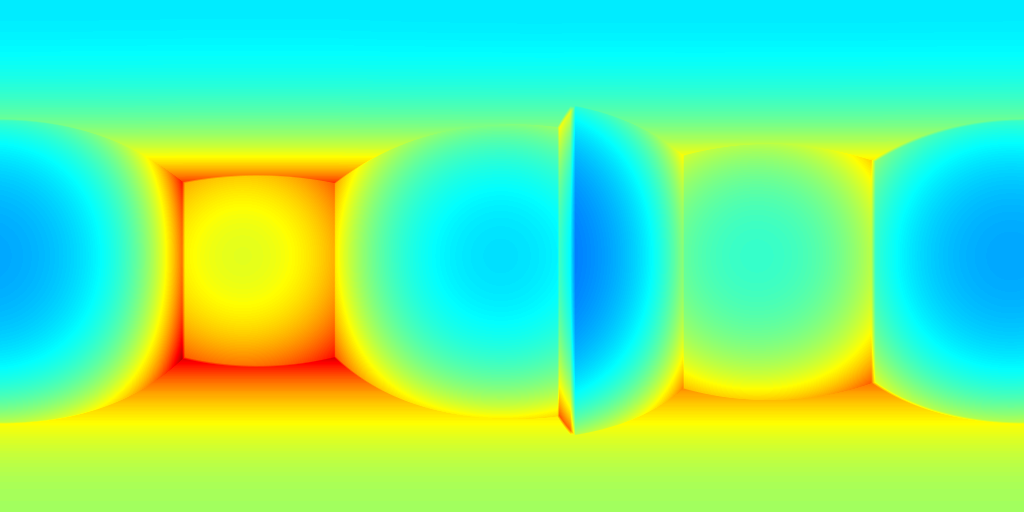}
\includegraphics[width=\linewidth]{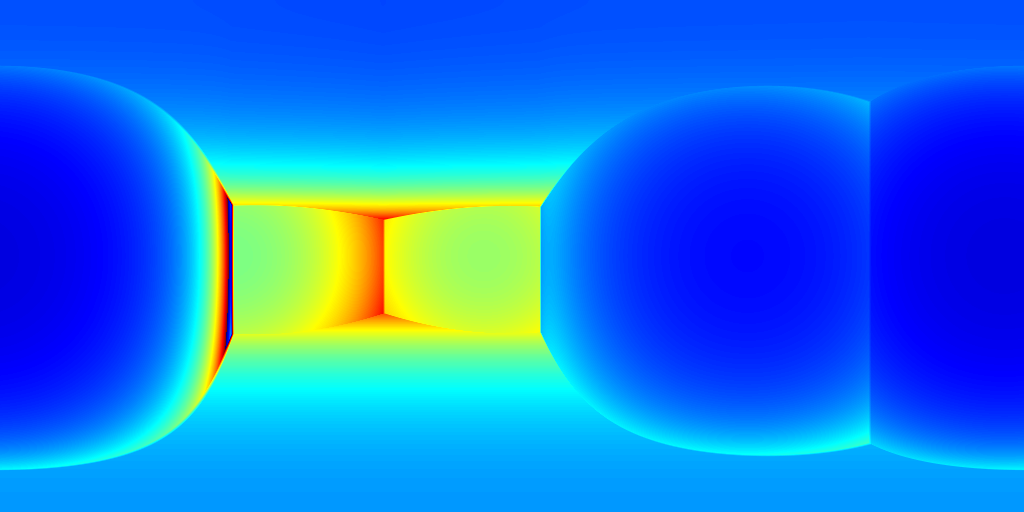}
(c) HorizonNet+SP
\end{minipage}
\begin{minipage}{0.24\linewidth}
\centering
\includegraphics[width=\linewidth]{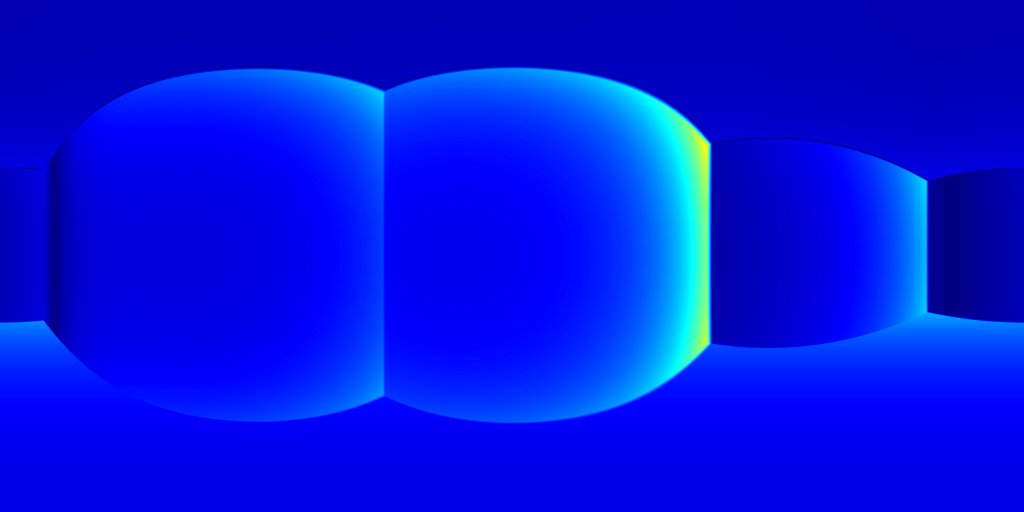}
\includegraphics[width=\linewidth]{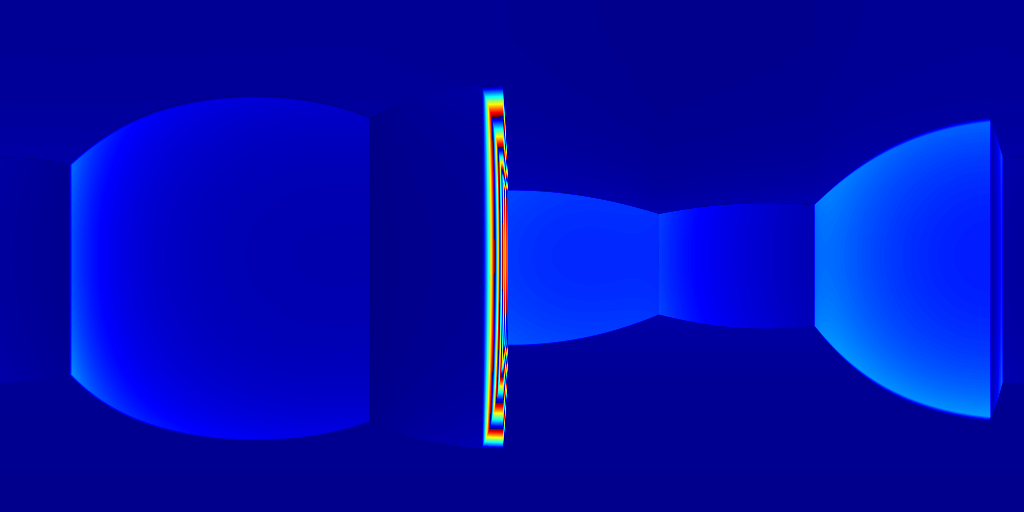}
\includegraphics[width=\linewidth]{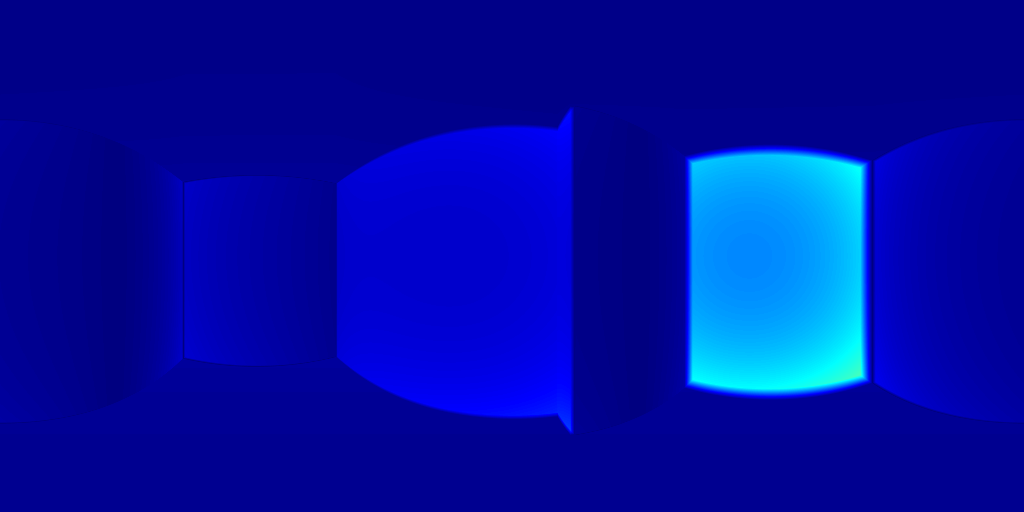}
\includegraphics[width=\linewidth]{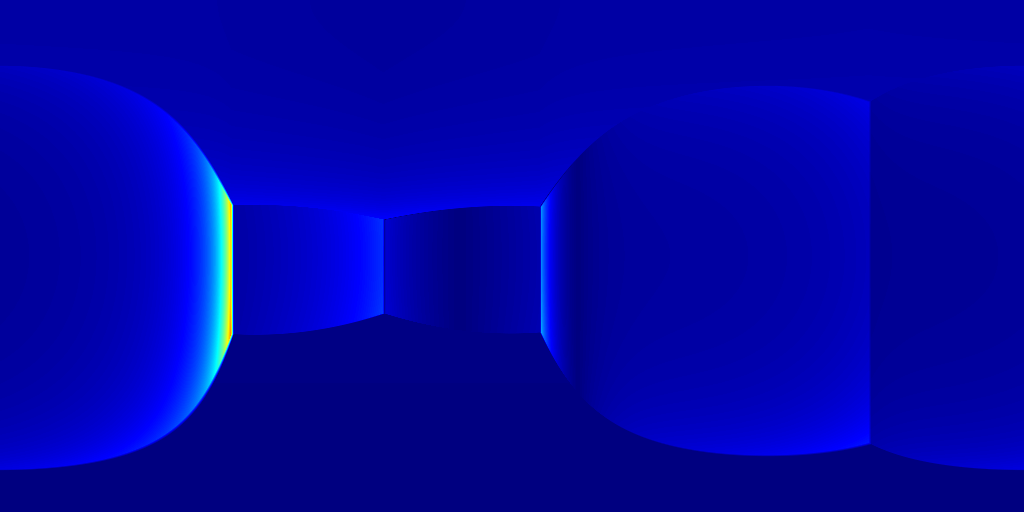}
(d) Ours
\end{minipage}
\caption{The qualitative results of layout boundary and depth error map on the ZInD dataset. It shows that accurate layout boundaries and depth maps can still be reconstructed in unfurnished and complex indoor scenes.}
\label{fig:zind} 
\end{figure*}

For ZInD dataset, we show even more salient improvement compared to the state-of-the-art methods as shown in Table~\ref{tab:Accuracy_zind}. This dataset contains more unfurnished and extremely texture-less regions. Such challenging environments highlight the robustness of our network against monocular prediction. We achieve an improvement of 20.6\% in terms of depth RMSE, and 64.3\% in the scale error. Compared to experiments on 2D-3D-S, our method shows a bigger improvement in terms of coherency on ZInD with a wider range of heights the images are captured.
We visualize the reconstructed layouts and the depth error maps in Figure~\ref{fig:zind} similar to what we do for 2D-3D-S. Note that even when 3D layouts from existing methods are far from ground truth as shown in the first row, our method still derives accurate layout depth from multi-view information. Such a challenging dataset suggests that our method can reconstruct the 3D layouts with much higher quality.  
\begin{table}
\centering
\small
\begin{tabular}{|c|c|c|c|}
\hline
ZInD & Depth RMSE & Scale error & Coherency \\
\hline
HorizonNet & 0.45 & 0.27 & 0.25 \\
\hline
HorizonNet+SP & 0.34 & 0.14 & 0.23 \\
\hline
HoHoNet+SP & 0.35 & 0.14 & 0.23 \\
\hline
Ours & \textbf{0.27} & \textbf{0.05} & \textbf{0.18} \\
\hline
\end{tabular}
\caption{Evaluation for 3D layout reconstruction on ZInD dataset. Our method performs the best in all metrics, especially for the scale error.}
\label{tab:Accuracy_zind}
\end{table}

\par

\begin{figure*}
\includegraphics[width=\linewidth]{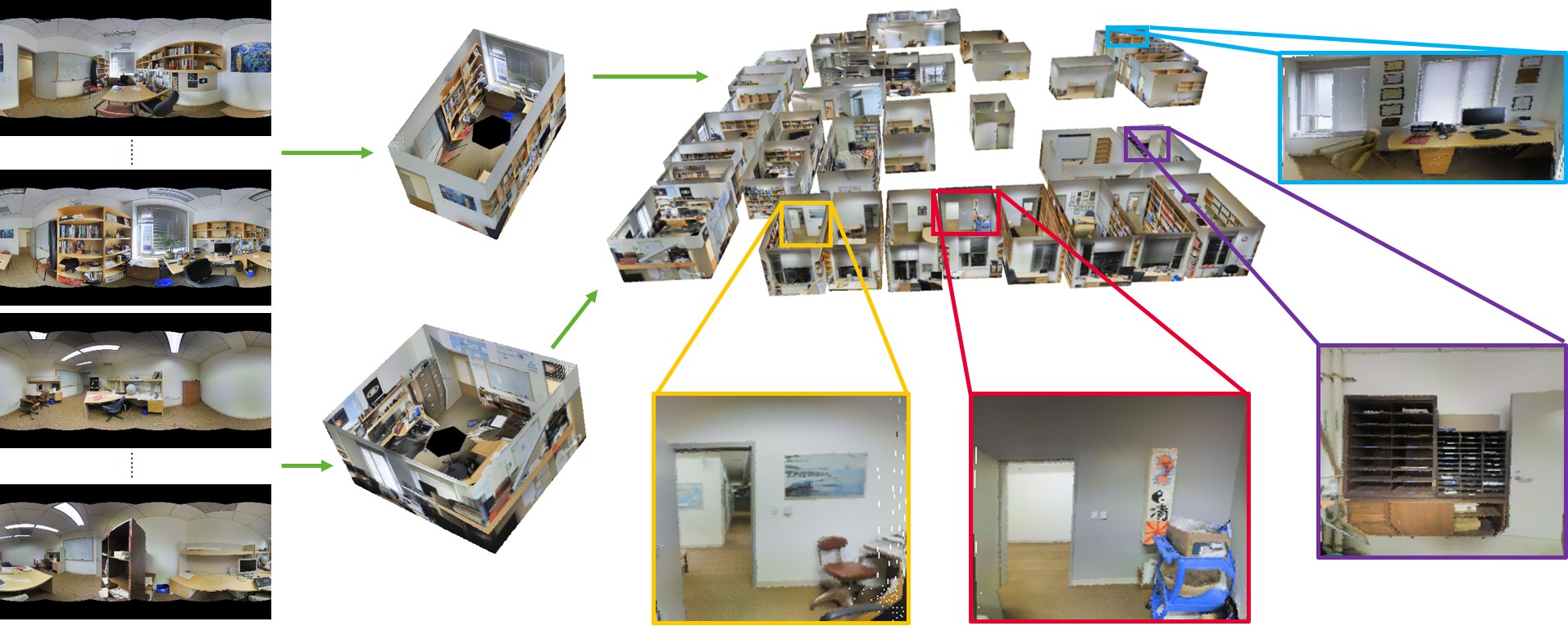}
\caption{The scene-level reconstruction results on 2D-3D-S dataset. The details of the reconstructed 3D layouts are zoomed-in. We carefully check and ensure that layout reconstruction is reasonable almost everywhere in this scene.}
\label{fig:scene-results}
\end{figure*}

\subsection{Ablation study}
\label{sec:exp-ablate}
\paragraph{Semantic information}
While semantic information is important to layout estimation, there are potentially multiple solutions to introduce them into our system.
The most straightforward way which we suggest is to directly specify the confidence of regions according to predicted semantics. Specifically, we treat ceilings, floors, and walls as layout structures and set their confidence $c^s$ as one and others zero. An alternative solution is to learn the confidence from semantic features, in which case we pass semantic features through an MLP layer to obtain $c^s$. We compare these two choices with a version where $c^s$ is always set to one. We set $c^a=1$ to focus on semantic confidence and keep other settings the same as discussed in Section~\ref{sec:approach}. Table~\ref{tab:Ablation_semantic} shows evaluation of different choices of using semantics on 2D-3D-S dataset. We find both learned semantics or user-specified semantic confidence helps improve the accuracy, where our user-specified version performs the best. Another interesting finding is that coherency is nearly the same among all methods, suggesting that our MVS module robustly produces coherent geometry no matter whether cluttered regions are taken into consideration.
\begin{table}
\centering
\small
\begin{tabular}{|c|c|c|c|}
\hline
2D-3D-S & Depth RMSE & Scale error & Coherency \\
\hline
Without sem & 0.24 & 0.08 & 0.10 \\
\hline
Learned sem & 0.22 & 0.07 & 0.10 \\
\hline
User spec sem & \textbf{0.21} & \textbf{0.06} & 0.10 \\
\hline
\end{tabular}
\caption{Results for alternative solutions of using semantics on 2D-3D-S dataset. Semantics can be ignored, learned by the network, or specified by users. While both ways of using semantics improve the results, explicitly setting confidence according predicted semantics performs the best.}
\label{tab:Ablation_semantic}
\end{table}

\paragraph{Self-attention confidence}
Besides semantics information, we add additional degrees of freedom for the network to adjust confidence with a self-attention module.
We aim to study the effectiveness of the self-attention confidence $c^a$ and its relationship to semantic confidence $c^s$. In Table~\ref{tab:Ablation_self_attention}, we evaluate a vanilla version by setting both confidence to one, a version where $c^s=1$ and $c^a$ are learned by our network, and a full version where both $c^s$ and $c^a$ are activated. We find attention helps to improve the performance, and the full version achieves the best performance. This suggests that attention is beneficial and should be jointly considered with semantic information.
\begin{table}
\centering
\small
\begin{tabular}{|c|c|c|c|}
\hline
2D-3D-S & Depth RMSE & Scale error & Coherency \\
\hline
Fix $c^s$ and $c^a$ & 0.24 & 0.08 & 0.10 \\
\hline
Activate $c^a$ & 0.21 & 0.06 & 0.09 \\
\hline
Activate $c^s$ and $c^a$ & \textbf{0.18} & \textbf{0.05} & \textbf{0.09} \\
\hline
\end{tabular}
\caption{Evaluation on the influence of self-attention and relationship to semantic confidence. Three conditions are considered. The first row fix both semantic and attention confidence. The second row only fix semantic confidence. The third row activates both confidence.}
\label{tab:Ablation_self_attention}
\end{table}

\paragraph{MVS with postprocessing}
An important question is whether it is feasible to derive a clean and accurate 3D layout by post-processing the MVS results. We design an experiment where MVS is processed with an orientation-fixed plane fitting to obtain the layout, where orientation is produced from our 2D layout network. We evaluate results from MVSNet~\cite{MVSNet}, post-processed by plane fitting (PF), and our network by comparing them with the ground truth depth map. Detailed statistics are shown in  Table~\ref{tab:Ablation_MVS}.
While it is possible to obtain a clean 3D layout from MVS, the accuracy is not satisfying. Specifically, the depth RMSE is similar to the monocular prediction from a monocular view with a scale prediction (e.g. Table~\ref{tab:Accuracy_2d3ds}, HorizonNet+SP). In contrast, our direct layout-based MVS module can significantly improve the 3D accuracy.
\begin{table}
\centering
\small
\begin{tabular}{|c|c|c|c|}
\hline
2D-3D-S & Depth RMSE & Scale error & Coherency \\
\hline
MVSNet & 0.24 & - & 0.10 \\
\hline
MVSNet+PF & 0.22 & 0.06 & 0.09 \\
\hline
Ours & \textbf{0.18} & \textbf{0.05} & \textbf{0.09} \\
\hline
\end{tabular}
\caption{Ablation experiments for the dense MVS methods on 2D-3D-S dataset. MVSNet is utilized as the MVS baseline and a plane fitting (PF) is further applied to extract the layout.}
\label{tab:Ablation_MVS}
\end{table}

\subsection{Scene-level Reconstruction}
\label{sec:exp-result}
Finally, we demonstrate a holistic reconstruction of the entire 2D-3D-S dataset given thousands of panoramas. Due to accurate view-based 3D layout predictions, we are able to apply layout fusion (Section~\ref{sec:approach-fusion}) and obtain globally consistent layout geometry of the scene. As shown in Figure~\ref{fig:scene-results}, we obtain coherent and reasonable structure layouts almost everywhere in such a big building, opening the potential for robust and abstract reconstruction of large scenes from pure RGB cameras.

\section{Conclusion}
We present a novel end-to-end framework for 3D layout reconstruction with multi-view panoramas. We seamlessly combine monocular layout estimation and an MVS network to obtain an accurate 3D layout of scenes in a large scale. Technically, we build a layout cost volume from layout homography, compress the volume into a 1D probability map and regress a single depth value for each layout element. Our solution significantly improves the geometry accuracy of the reconstructed layout.

%%%%%%%%% REFERENCES
{\small
\bibliographystyle{ieee_fullname}
\bibliography{PaperForReview}
}

\end{document}